\documentclass[cameraready]{Interspeech}

\title{SPARCLE: SPeaker-aware Aligned Representations via Contrastive Language Embeddings}

\author[affiliation={1,2}, equalcontribution, correspondingauthor]{Priyam}{Mazumdar}
\author[affiliation={1,2}, equalcontribution, correspondingauthor]{Yurii}{Halychanskyi}
\author[affiliation={1}]{Steven}{Guo}
\author[affiliation={1}]{Mark}{Hasegawa-Johnson}
\author[affiliation={1,2}]{Volodymyr}{Kindratenko}

\address{
    $^1$ University of Illinois Urbana-Champaign, USA \\
    $^2$ National Center for Supercomputing Applications, USA
}

\email{priyamm2@university.edu, yuriih2@illinois.edu}


\usepackage{comment}
\usepackage{multirow}
\usepackage{booktabs}

\begin{document}

\maketitle

\begin{abstract}


Recent advances in speech synthesis have shifted from phoneme representations to direct grapheme modeling. While phonemes address the one-to-many mapping between text and acoustics, they rely on grapheme-to-phoneme (G2P) systems that fail to capture speaker-specific acoustic variation. Prior work demonstrates that grapheme-based models outperform phoneme-based systems at scale, but not in low-resource settings. 

In this paper, we propose SPARCLE, a speaker-aware grapheme representation model that enriches characters with their precise acoustic realizations. SPARCLE is trained with a contrastive objective to align graphemes with corresponding Wav2Vec2 acoustic representations while conditioned on speaker identity. The resulting model serves as a replacement to G2P systems for downstream text-to-speech (TTS) tasks. We demonstrate that SPARCLE improves generation quality, reducing word error rates by half in extreme low-resource settings compared to standard grapheme-based models. 

\end{abstract}

\section{Introduction}

Generative modeling has recently achieved remarkable progress across multiple modalities~\cite{ramesh2022hierarchical,saharia2022photorealistic,brown2020language,agostinelli2023musiclm}, with speech synthesis benefiting substantially~\cite{chen2024f5tts,wang2023valle}. Phonemes have long been a popular input choice, as they explicitly encode pronunciation and mitigate the one-to-many mapping problem where a single grapheme sequence can yield multiple acoustic realizations~\cite{cheng2024survey}. Using a standardized phone inventory such as the International Phonetic Alphabet (IPA) also facilitates multilingual synthesis by avoiding the need to manage separate grapheme sets across languages~\cite{adams2019massively}.

However, phoneme-based approaches rely on grapheme-to-phoneme (G2P) conversion, which requires accent- and dialect-aware models. Training such G2P systems typically depends on phoneme labels that are expensive to obtain, or handcrafted rules and pronunciation dictionaries that are costly to develop and difficult to scale~\cite{cheng2024survey}. At the same time, recent studies show that when sufficient training data is available, character-based models can match or even surpass phoneme-based systems in synthesis quality~\cite{kharitonov2023speakreadprompthighfidelity}, motivating modern text-to-speech (TTS) frameworks such as F5-TTS~\cite{chen2024f5tts} to adopt graphemes directly. 

In this paper, we propose a model that enriches grapheme inputs with fine-grained, context-specific embeddings that capture their exact acoustic realizations, while conditioned on speaker embeddings derived from the corresponding speech.

\begin{figure}[t]
    \setlength{\abovecaptionskip}{-2pt}
    \setlength{\belowcaptionskip}{-12pt}
    \centering
    \includegraphics[width=\linewidth]{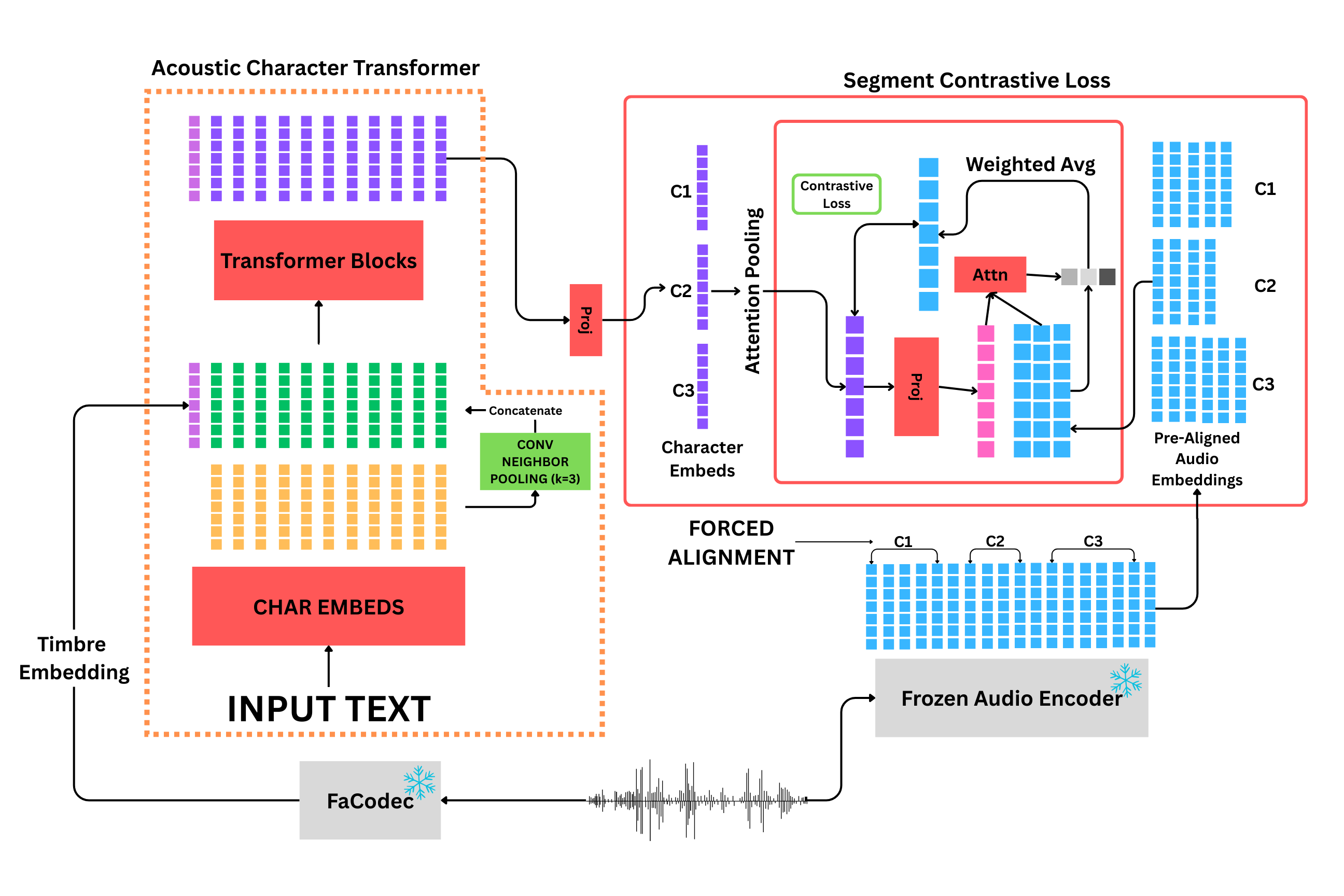}
    \caption{SPARCLE Architecture}
    \label{fig:sparcle_architecture}
\end{figure}

\section{Related Work}

Contrastive Learning has emerged as a powerful paradigm for learning joint embedding spaces across different modalities. The main task is to directly optimize representations such that semantically corresponding inputs from different modalities are mapped close together. One influential work from which we take inspiration from is Contrastive Language-Audio Pretraining (CLAP) \cite{clap} that learns joint audio-text representations.

CLAP has two limitations for text-encoding 1) CLAP relies on paired audio-text data where the textual modality consists of captions, tags or semantic descriptors rather than exact linguistic transcriptions. 2) CLAP performs contrastive learning at the sequence level, aligning the entire audio clip with the corresponding text descriptions rather than learning more fine temporal or linguistic granularity of characters. 

A variety of Neural G2P models have been proposed, including ByT5-based Multilingual G2P system \cite{byt5g2p} trained across 100 languages. Due to the scarcity of aligned text-phoneme corpora, these approaches rely on pronunciation dictionaries that provide mappings between words and their IPA transcriptions. Consequently, training is performed at the word level rather than the utterance level. This limits the models ability to capture phonetic variation arising from contextual effects such as co-articulation \cite{coarticulation}, prosody, and speaker-specific pronunciation, all of which depend on the broader utterance context rather than isolated words. 

\section{Method}

The main limitation of using graphemes in TTS systems is pronunciation ambiguity \cite{Han2024}. Many languages, such as as English, have inconsistent spelling-to-sound mappings. For example, the word \textit{read} can be phonetically spelled as \textipa{/ri:d/} or \textipa{/rE:d/}, depending on the tense even with the same grapheme sequence. This inconsistency motivates the use of G2P conversion as an input representation for TTS systems; however, this approach often overlooks the fact that G2P rules are frequently dialect specific \cite{phonetic_transformations}.

Our proposal is to inject acoustic information directly into character-level (grapheme-level) models through contrastive pretraining. By combining our grapheme inputs with acoustic cues, the model will learn context-sensitive pronunciation patterns without explicit phoneme annotations. This also allows us to leverage large datasets of high-quality grapheme-transcribed speech. 

\subsection{Character Alignment}

For extracting acoustic embeddings, we leveraged the Wav2Vec2 \cite{baevski2020wav2vec20frameworkselfsupervised} family of models. To enable character-level contrastive learning, it was necessary to obtain embeddings from the acoustic model that corresponded to each letter in the transcript. To achieve this, we force-aligned each transcript to its corresponding audio, recording the indices of the acoustic embeddings associated with each character. The end result was the mapping between each letter of our transcript and the corresponding set of embedding vectors from Wav2Vec2. 

This operation was performed across the entire Librispeech-960h \cite{librispeech} dataset, creating a rich set of character-acoustic pairs spanning thousands of utterances and a large range of speakers.

\subsection{Speaker Identity}

To create a speaker-aware model, we require embedding vectors that capture the speaker identity. We avoid learning a speaker-specific embedding matrix, as this would hinder generalization to downstream tasks involving previously unseen speakers. Instead, we adopt the FaCodec timbre embeddings \cite{facodec} for two reasons. 1) We found that FaCodec timbre embeddings provide strong separation between speakers, even in the presence of accented speech. 2) These embeddings are extracted from a pretrained, speaker agnostic model, and they enable robust conditioning on speaker identity without retraining or adaptation when encountering new speakers, at least for the English language. Additionally, the FaCodec model can be fully trained with audio samples only \cite{zhang2023amphion} making it easy to adapt this architecture to new dialects and languages for future work

\subsection{Character Transformer}

The character transformer, illustrated in figure \ref{fig:sparcle_architecture}, operates as a standard encoder-style transformer trained at the character level, with tokens restricted to uppercase letters A–Z and a space character. Although the transformer captures global contextual information through self-attention, we hypothesize that the pronunciation of a character depends heavily on its local neighbors. For instance, how a letter is articulated can change depending on the letters surrounding it. To explicitly model this neighboring context, we apply a 1D convolution with kernel size 3 over the the character embeddings. The resulting short-context embeddings are then concatenated with the original character embedding and projected to the transformers input dimension. 

To condition the model on speaker identity, the speaker embedding is pre-pended to the character embeddings before being processed by the transformer encoder. Because FaCodec timbre embeddings exhibit a high variance, we applied unit normalization prior to concatenation. Additionally, random dropout is applied with $p=0.5$ to mitigate overfitting to the set of speakers present in the training data. After encoding, the speaker token is discarded, and the resulting sequence is used to compute the contrastive loss objective. 

The architecture followed the standard base transformer shape of 12 blocks each with 12 heads of attention and an embedding dimension of 768. All character embeddings had a dimension of 128, and the neighbor convolutions computed an additional 128 dimensional embeddings. These were concatenated into 256 dimensional feature vectors that were then projected to the transformers embedding dimension. 

\subsection{One-To-Many Contrastive Pre-Training}

\begin{table*}[t]
\centering
\small
\caption{WER and EER comparison across dataset durations. Char Embeddings is the ParrotTTS using standard character embeddings, Phone Embeddings is ParrotTTS using phoneme embeddings from the g2p\_en phonemizer, and all other models use our SPARCLE architecture with various numbers of layers unfrozen during training (indicated by K). Comparison of SPARCLE with and without included Timbre embeddings from FaCodec is identified as +/- T respectively}
\begin{tabular}{ll|cc|cc|cc|cc|cc}
\toprule

& & \multicolumn{2}{c|}{10m} 
  & \multicolumn{2}{c|}{30m} 
  & \multicolumn{2}{c|}{1H} 
  & \multicolumn{2}{c|}{5H} 
  & \multicolumn{2}{c}{10H} \\

\cmidrule(lr){3-4}
\cmidrule(lr){5-6}
\cmidrule(lr){7-8}
\cmidrule(lr){9-10}
\cmidrule(lr){11-12}

& & -T & +T 
  & -T & +T
  & -T & +T
  & -T & +T
  & -T & +T \\

\midrule

\multirow{6}{*}{WER (\%)}
& Char Embeddings & 85.7 & - & 35.3 & - & 24.7 & - & 17.2 & - & 14.4 & - \\
& Phone Embeddings & 96.0 & - & 32.6 & - & 26.8 & - & 27.7 & - & 25.1 & - \\
& Frozen SPARCLE    & 53.4 & 69.2 & 16.3 & 20.7 & 13.7 & 18.0 & 13.2 & 13.3 & 13.7 & 14.8 \\
& K=1 SPARCLE       & 47.1 & 58.7 & 14.7 & 17.8 & 12.4 & 14.3 & 12.2 & 12.6 & 12.0 & 11.6 \\
& K=3 SPARCLE       & 43.5 & 42.2 & 12.3 & 11.5 & 9.3 & 9.2 & 11.7 & 10.2 & 11.4 & 10.6 \\
& K=7 SPARCLE       & 42.4 & \textbf{42.2} & 12.3 & \textbf{10.0} & 9.2 & \textbf{7.5} & 11.0 & \textbf{8.5} & 11.4 & 11.0 \\
& Unfrozen SPARCLE  & 52.1 & 49.3 & 14.9 & 12.8 & 11.6 & 9.1 & 11.9 & 8.9 & 11.3 & \textbf{9.9} \\

\midrule

\multirow{6}{*}{EER (\%)}
& Char Embeddings & 2.93 & - & 1.95 & - & 1.93 & - & 1.37 & - & 1.17 & - \\
& Phone Embeddings & 3.32 & - & 1.76 & - & 1.59 & - & 1.37 & - & 1.33 & - \\
& Frozen SPARCLE    & 2.10 & 2.15 & 1.37 & 1.56 & 1.02 & 1.21 & 1.33 & 1.33 & 1.56 & 1.00 \\
& K=1 SPARCLE       & 2.34 & 2.39 & 1.71 & 1.56 & 1.76 & 1.37 & 1.00 & 0.98 & 0.97 & 1.02 \\
& K=3 SPARCLE       & 1.95 & 1.99 & 1.54 & 1.56 & 1.37 & 1.18 & 0.98 & 1.18 & 0.98 & 1.38 \\
& K=7 SPARCLE       & \textbf{1.95} & 2.00 & 1.52 & \textbf{1.19} & 1.56 & \textbf{1.13} & 1.21 & \textbf{0.98} & 1.17 & \textbf{0.76} \\
& Unfrozen SPARCLE  & 2.19 & 2.14 & 1.52 & 1.59 & 1.34 & 1.17 & 0.97 & 0.98 & 0.94 & 0.83 \\

\bottomrule
\end{tabular}
\label{wer_eer_results}
\end{table*}

We leverage contrastive learning to align our character embeddings with their corresponding acoustic embeddings.  However, the mapping between characters and acoustic frames is inherently one-to-many: on average, each character token A–Z corresponds to approximately 2.8 Wav2Vec2 embeddings, while each space character corresponds to about 6.7 embeddings. To convert this variable-length mapping into a single, fixed-size target per character, we employ an attention pooling mechanism over the Wav2Vec2 embeddings. In this setup, the Wav2Vec2 model is kept frozen with no trainable parameters.

For each character, we compute attention weights over its associated acoustic embeddings using a learned projection of the character embeddings. These weights allow the model to extract the most relevant acoustic frames, producing a pooled audio embedding. The character and pooled acoustic embedding are then L2 Normalized and the cosine similarity is computed between every pair, producing $N$ targets and $N^2 - N$ negatives. We use use a constant temperature parameter of 0.1 to scale the logits before the contrastive loss is computed. 

\subsection{Pre-Training Details}

All models were trained for 200K steps and the best checkpoint on the test data was taken. They were trained with a batch size of 1024 transcript/audio pairs on 4xGH200 GPUs. We use a learning rate of 1e-4 that is Cosine decayed through all of training as well as weight decay of 0.1. Lastly, any audio shorter than 2 seconds or longer than 20 seconds were removed from the dataset.




\section{Experiment Details}

\begin{table}[t]
  \caption{Effect of SPARCLE relative to character baselines at 1 hour of VCTK training data. Results are reported separately for each TTS backend.}
  \label{tab:_1h_baseline}
  \centering
  \setlength{\tabcolsep}{3pt}
  \begin{tabular}{ l l r r }
    \toprule
    \textbf{TTS Backend} & \textbf{Representation} & \textbf{WER (\%)} & \textbf{EER (\%)} \\
    \midrule
    \multirow{2}{*}{ParrotTTS}
      & Characters (baseline) & 24.8 & 1.9 \\
      &  (ours)          & \textbf{9.2} & \textbf{1.5} \\
    \midrule
    \multirow{2}{*}{VITS}
      & Characters (baseline) & 121.7 & 5.6 \\
      &  (ours)          & \textbf{117.34} & \textbf{3.9} \\
    \bottomrule
  \end{tabular}
\end{table}

We evaluate SPARCLE as a drop-in replacement for character embeddings in two TTS backends: (i) ParrotTTS~\cite{parrottts}, a modular system that predicts discrete self-supervised units from text and synthesizes waveforms with a separate neural vocoder; and (ii) VITS~\cite{vits}, an end-to-end TTS model. Our goal is to assess whether \emph{speaker-aware, acoustically grounded character embeddings} improve pronunciation accuracy and speaker consistency, particularly in low-resource multi-speaker training.

\subsection{Data}

\textbf{SPARCLE Pre-Training.} SPARCLE is pre-trained on LibriSpeech-960h~\cite{librispeech}. Forced alignment is used to associate transcript characters with acoustic frames, and SPARCLE is trained to align character embeddings with frozen Wav2Vec2 representations~\cite{baevski2020wav2vec20frameworkselfsupervised}. Unless stated otherwise, we use Wav2Vec2-Large acoustic targets from layer $-17$ (8th hidden representation), which we found empirically to provide the strongest phonetic supervision among the layers we tested.

\noindent\textbf{Downstream TTS.} Downstream synthesis is trained and evaluated on VCTK v0.92~\cite{vctk} using microphone \texttt{mic2}, with 108 speakers. We construct speaker-balanced low-resource training subsets while keeping the speaker set fixed across all budgets. We consider training budgets of 10 minutes, 30 minutes, 1 hour, 5 hours, and 10 hours, corresponding to approximately
(10m) $\sim$1 utt/spk,
(30m) $\sim$5 utt/spk,
(1h) 9 utt/spk,
(5h) 45 utt/spk,
(10h) 91 utt/spk.
All configurations are evaluated on a \emph{fixed} held-out test set of 512 utterances sampled across speakers; this same test set is used for both backends and all training budgets.

VCTK was selected to explicitly evaluate the model's robustness to domain shift. The SPARCLE model was pretrained on LibriSpeech, which is predominantly composed of US English speakers \cite{globe}. In contrast, VCTK is dominated by British English speakers. This introduces a natural domain mismatch between the pretraining and downstream finetuning data, and allows us to assess how well SPARCLE generalized beyond the distribution it was originally trained on. 

\subsection{Backends and SPARCLE integration}

\textbf{ParrotTTS.} We replace ParrotTTS’s standard character embedding layer with SPARCLE outputs, followed by a learned linear projection to the ParrotTTS model dimension. Non A--Z symbols (e.g., punctuation) are handled by a small auxiliary embedding table, which is randomly initialized and trained jointly with the downstream TTS model.

ParrotTTS relies on an aligner trained in an ASR-style manner. In our setup, the aligner operates directly on character sequences and produces character-level duration estimates. These durations are used as ground-truth supervision for the downstream duration predictor. Both the aligner and the vocoder are kept fixed across all experiments: they are trained once on the full VCTK training set and reused for all low-resource budgets, ensuring that observed differences are attributable solely to changes in the linguistic representation.

\noindent\textbf{VITS.} We integrate SPARCLE analogously: SPARCLE replaces the initial character embedding lookup, and its outputs are projected to the VITS text-encoder hidden size. The remainder of the VITS architecture (text encoder stack, duration modeling, flow, and decoder) is unchanged; SPARCLE only alters the input representation.

\subsection{Fine-tuning regimes and speaker conditioning}

We evaluate SPARCLE along two axes: (i) how much of SPARCLE is adapted during downstream TTS training, and (ii) whether SPARCLE is conditioned on speaker identity.

\noindent\textbf{Adaptation depth.} In all cases, SPARCLE is initialized from the LibriSpeech pre-trained checkpoint. We vary which transformer blocks are trainable:
\begin{itemize}
    \item \textbf{Frozen}: SPARCLE is used as a fixed feature extractor.
    \item \textbf{Partial fine-tuning} ($K \in \{1,3,7\}$): we unfreeze the \emph{last} $K$ transformer blocks and train them jointly with the downstream TTS model.
    \item \textbf{Fully unfrozen}: all SPARCLE layers are jointly fine-tuned with downstream training.
\end{itemize}
This setup probes the tradeoff between adapting SPARCLE to the downstream domain and preserving its pre-training signal.

\noindent\textbf{Speaker conditioning.} We report results with and without timbre conditioning. For timbre conditioning, we prepend a special speaker token whose value is the FaCodec timbre embedding~\cite{facodec} extracted from the target speaker. Dropout is applied to the timbre embedding to reduce overfitting.

\subsection{Downstream training details}

\textbf{ParrotTTS training.} The text-to-units module is trained with AdamW (initial learning rate $1\times10^{-4}$, $\beta=(0.9,0.98)$, weight decay 0), linear warmup for 2k steps, and 50k total steps, with gradient clipping at 1.0 and batch size 6.

\noindent\textbf{VITS training.} VITS is trained with AdamW (learning rate $2\times10^{-4}$, $\beta=(0.8,0.99)$, $\epsilon=10^{-9}$, batch size 64).

\subsection{Evaluation metrics}

We report pronunciation accuracy via word error rate (WER) and speaker consistency via equal error rate (EER).

\textbf{WER.} Generated speech for the 512-utterance test set is transcribed using Whisper~\cite{whisper} (\texttt{small}). Transcripts are normalized (lower-casing, punctuation removal, and digit normalization), and WER is reported as a micro-average over the full test set.

\textbf{EER.} Speaker consistency is measured using a pretrained ECAPA-TDNN \cite{ecapatdnn} speaker verification model from SpeechBrain, specifically the VoxCeleb-trained checkpoint \texttt{speechbrain/spkrec-ecapa-voxceleb}. Genuine trials compare generated utterances with their corresponding reference utterances from the same speaker, while impostor trials compare against reference utterances from other speakers. No score normalization is applied.

\section{Results}

\textbf{SPARCLE consistently improves low-resource pronunciation.}
Table~\ref{wer_eer_results} shows that replacing character embeddings with SPARCLE improves WER over the character-only baseline across all budgets. The gains are largest in the lowest-resource regimes. At 10 minutes, the character baseline WER is 85.7\%, while SPARCLE reduces WER to 42.2\% (timbre, $K{=}7$). At 1 hour, WER drops from 24.7\% to 7.5\% (timbre, $K{=}7$). As training data increases, the gap narrows, indicating that with sufficient supervised data the TTS backend can learn more of the grapheme-to-acoustics mapping directly.

\textbf{Poor G2P Phonemizer Alignment}
To benchmark our architecture against a standard G2P module, we use the g2pE phonemizer \cite{g2pE2019} to convert VCTK transcripts to phoneme sequences. As shown in Table~\ref{wer_eer_results}, phoneme-based models underperform at all training budgets except 30 minutes. This contrasts with Kharitonov et al. \cite{kharitonov2023speakreadprompthighfidelity}, who found phonemes effective at low resources. However, their experiments used LJSpeech, which, like the CMU Pronunciation Dictionary underlying g2pE, represents North American English. VCTK is primarily British English, making this phonemizer a poor fit. This mismatch underscores the importance of pairing pronunciation dictionaries with the target speaker domain, and motivates our move away from them entirely.

\textbf{Partial fine-tuning yields the best adaptation--forgetting tradeoff.}
Across budgets, unfreezing a small number of upper SPARCLE layers yields the best WER/EER tradeoff. For example, without timbre at 10 minutes, WER improves from 53.4 (Frozen) to 42.4 ($K{=}7$), while fully unfrozen SPARCLE degrades to 52.1. A similar pattern appears at 1 hour (best 9.2 at $K{=}7$; fully unfrozen 11.6). These trends are consistent with partial fine-tuning enabling adaptation to VCTK while avoiding excessive forgetting of the LibriSpeech pre-training signal.

\textbf{Speaker conditioning helps primarily when SPARCLE can adapt.}
Timbre conditioning is not uniformly beneficial. When SPARCLE is frozen, adding timbre often degrades WER (e.g., at 1 hour Frozen increases from 13.7\% to 18.0\%). In contrast, once SPARCLE is partially fine-tuned, timbre conditioning becomes beneficial and yields the best overall results. At 1 hour, $K{=}7$ improves from 9.2\% (no timbre) to 7.5\% (+timbre), and EER improves from 1.56\% to 1.13\%. These results indicate that speaker-aware character embeddings are effective when the conditioning pathway is calibrated to the downstream data distribution.

\textbf{Speaker consistency improves alongside pronunciation.}
EER follows similar trends to WER. SPARCLE improves EER relative to the character baseline across budgets, with the strongest gains observed under partial fine-tuning and timbre conditioning (e.g., at 10 hours: baseline 1.17 vs.\ 0.76 at $K{=}7$ +timbre), indicating improved speaker consistency in addition to intelligibility.

\textbf{Backend generality at 1 hour.}
As shown in Table~\ref{tab:_1h_baseline}, at 1 hour VITS exhibits very high WER in this multi-speaker low-resource setting (121.70\%). SPARCLE yields a modest WER reduction (117.39\%) while substantially improving EER (5.63 $\rightarrow$ 3.89). This suggests that in extreme low-resource regimes, end-to-end models may be dominated by data scarcity effects, limiting the impact of improved text representations, while speaker-consistency benefits remain observable.

\textbf{Sub-word Level Modeling}
We also experimented with subword-level modeling using the RoBERTa \cite{roberta} token dictionary in place of character inputs. The Wav2Vec2 embedding alignments were recomputed at the token level. The results were poor, yielding 94.4\% WER on LJSpeech with audibly muffled speech and a loss of all acoustic details. We attribute this to excessive temporal aggregation, as the RoBERTa dictionary has a median token length of 5 characters. No further experiments were conducted at the sub-word level.

\section{Conclusions and Future Work}

We have demonstrated that SPARCLE encodes sufficiently rich acoustic information to support high-quality downstream TTS, even in low-resource conditions. Remarkably, with as little as 30 minutes of audio across all speakers, the model produces coherent and intelligible speech. This suggests that SPARCLE captures robust, transferable representations that generalize well beyond the specific data distribution.

We envision three primary directions for extending this architecture. 1) Because SPARCLE requires only aligned audio-text pairs, it can be readily extended to multilingual speech encoding. 2) The speaker conditioning is implemented without explicit speaker ids. This allows speaker representations to be inferred directly from reference audio, speaker descriptors, or generated by an auxiliary network, enabling zero-shot speech generation of previously unseen speakers. 3) Modern TTS systems have shifted towards Speech Language Models such as VALL-E \cite{valle} which generate discrete speech tokens autoregressively, rather than relying on a separate duration predictor. SPARCLE embeddings are well aligned with this requirement and can replace traditional G2P modules with a learned alternative that preserves fine-grained grapheme-level acoustic cues.


\bibliographystyle{IEEEtran}
\bibliography{mybib}

\end{document}